\def\year{2021}\relax
\documentclass[letterpaper]{article} 
\usepackage{aaai21}  
\usepackage{times}  
\usepackage{helvet} 
\usepackage{courier}  
\usepackage[hyphens]{url}  
\usepackage{graphicx} 
\usepackage{graphicx}  
\usepackage{natbib}  
\usepackage{caption} 
\usepackage[switch]{lineno} 
\usepackage{enumerate}
\usepackage{makecell}
\usepackage{multirow}
\usepackage{booktabs,tabularx}
\usepackage{amsmath,amssymb}
\usepackage{algorithm}
\usepackage{algpseudocode}

\title{Scene Text Detection with Scribble Lines}
\author{

    Wenqing Zhang\textsuperscript{\rm 1},
    Yang Qiu\textsuperscript{\rm 1},
    Minghui Liao\textsuperscript{\rm 1},
    Rui Zhang\textsuperscript{\rm 2},
    Xiaolin Wei\textsuperscript{\rm 2},
    Xiang Bai\textsuperscript{\rm 1}
}
\affiliations{

    \textsuperscript{\rm 1}Huazhong University of Science and Technology,
    \textsuperscript{\rm 2}Meituan

    \{wenqingzhang, yqiu, mhliao, xbai\}@hust.edu.cn, \{zhangrui36, weixiaolin02\}@meituan.com
}

\begin{document}
\maketitle

\begin{abstract}

Scene text detection, which is one of the most popular topics in both academia and industry, can achieve remarkable performance with sufficient training data.
However, the annotation costs of scene text detection are huge with traditional labeling methods due to the various shapes of texts.
Thus, it is practical and insightful to study simpler labeling methods without harming the detection performance.
In this paper, we propose to annotate the texts by scribble lines instead of polygons for text detection. It is a general labeling method for texts with various shapes and requires low labeling costs.
Furthermore, a weakly-supervised scene text detection framework is proposed to use the scribble lines for text detection. The experiments on several benchmarks show that the proposed method bridges the performance gap between the weakly labeling method and the original polygon-based labeling methods, with even better performance.
We will release the weak annotations of the benchmarks in our experiments and hope it will benefit the field of scene text detection to achieve better performance with simpler annotations.

\end{abstract}

\begin{figure}[tb]
\centering
\includegraphics[width=1.0\linewidth]{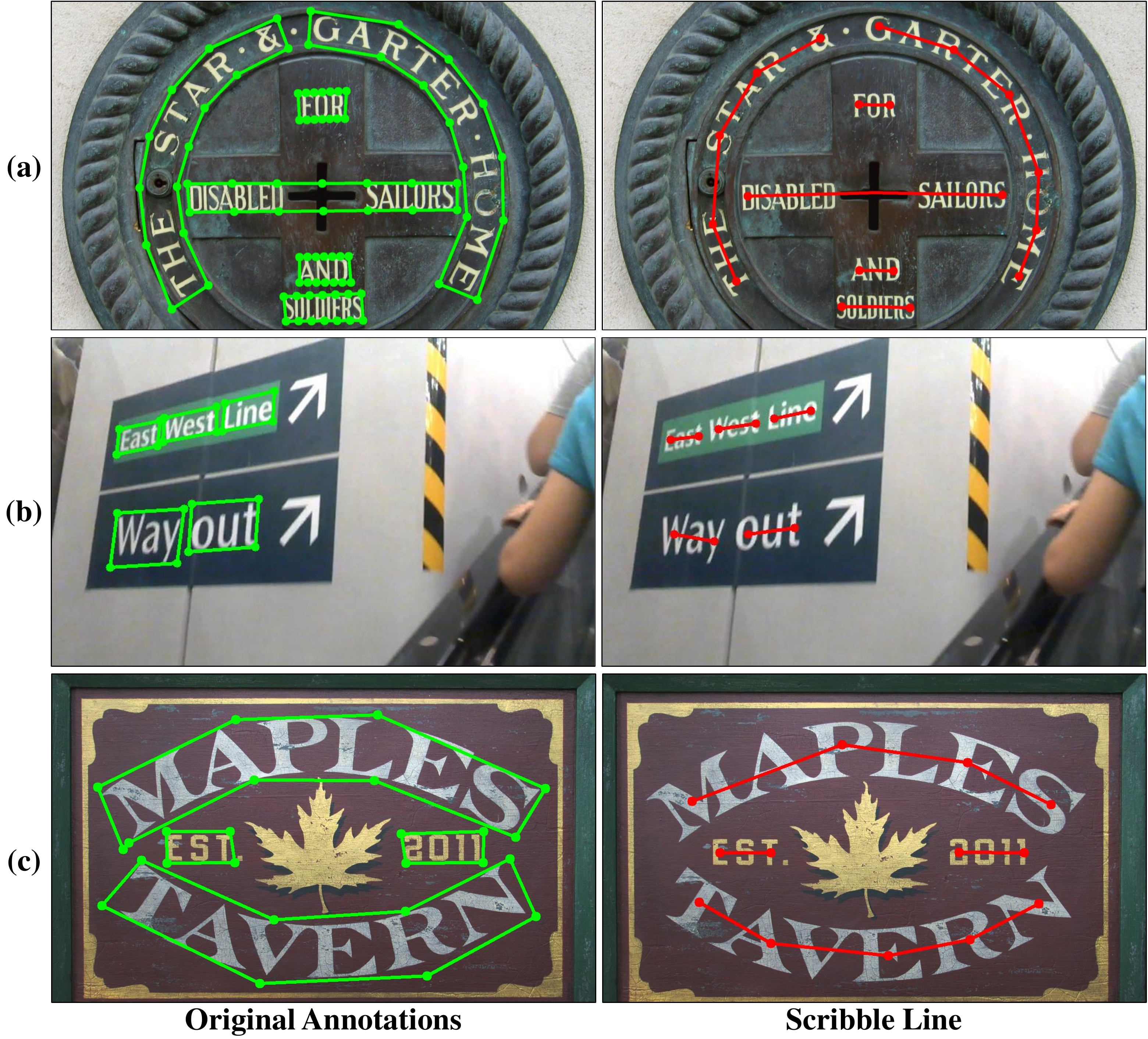}
\caption{Different annotations of the (a) CTW1500, (b) ICDAR 2015, and (c) Total-Text datasets. Green: original annotations; Red: our scribble line annotations.}
\label{fig:intro}
\end{figure}

\section{Introduction}
Scene text detection, which is a fundamental step for text reading, aims to locate the text instances in scene images. It has become one of the most popular research topics in academia and industry for a long time. In practice, detecting text in natural scene images is a basic task for various real-world applications, such as autonomous navigation, image/video understanding, and photo transcription.


Due to the variety of scene text shapes (\emph{e.g.} horizontal texts, multi-oriented texts, and curved texts), there is a high requirement for text location in scene text detection. 
In the early methods, they only focus on horizontal texts and adopt the axis-aligned bounding boxes to locate texts~\cite{ic13}, which is very similar to general object detection methods. 
In the following years, MSRA-TD500 dataset~\cite{TD500} and ICDAR 2015 dataset~\cite{ic15} adopt multi-oriented quadrilaterals to better locate the multi-oriented texts. 
Recently, in Total-Text dataset~\cite{totaltext} and CTW1500 dataset~\cite{ctw1500}, polygon annotations are used to outline texts with various shapes, and the Total-Text dataset even provides pixel-level text segmentation annotations.
Hence, for the general scene text detection task, accurate labeling is time-consuming and laborious.

\begin{figure}[tb]
\centering
\includegraphics[width=1.0\linewidth]{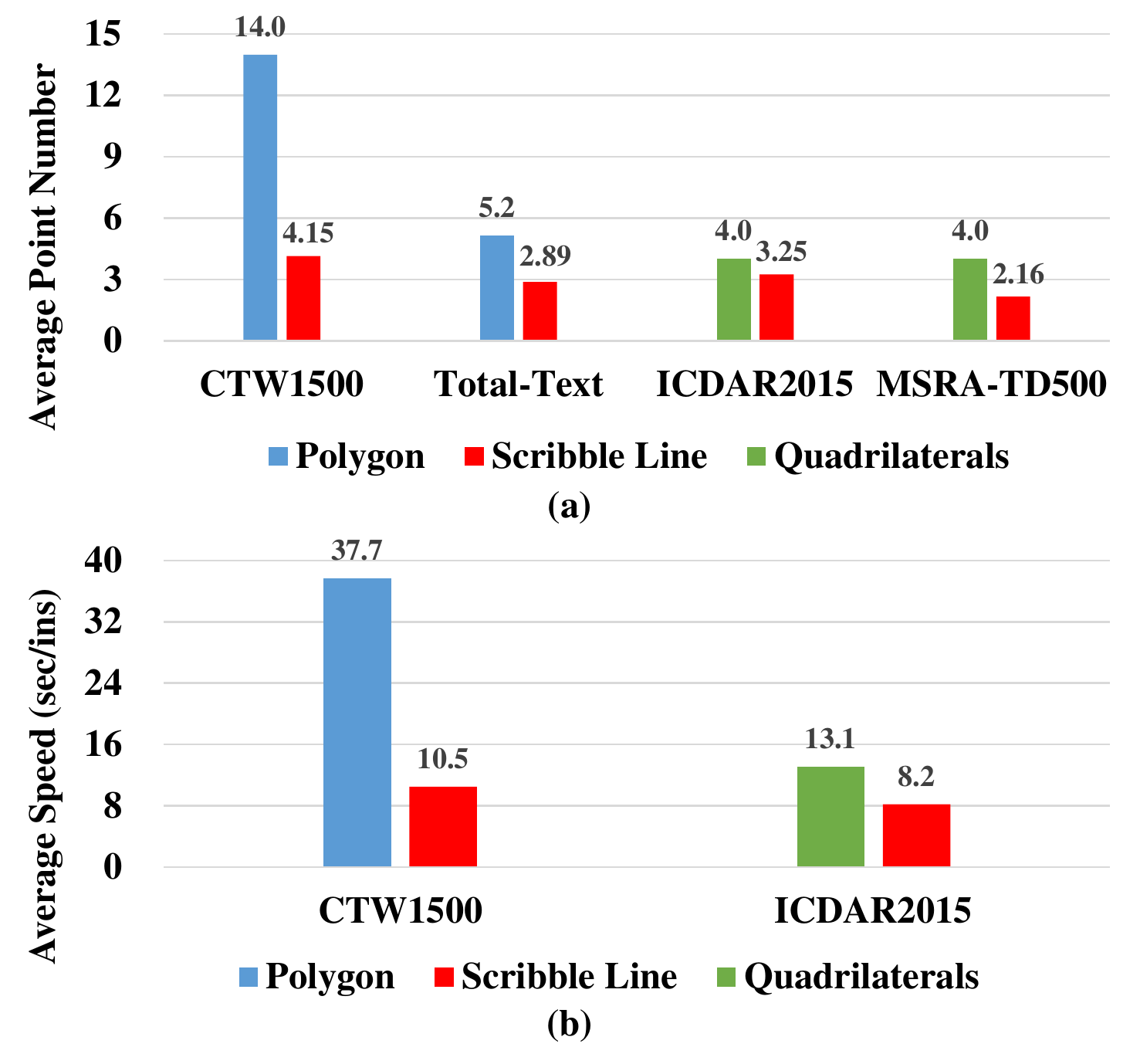}
\caption{Annotation cost comparison by (a) the average labeled points for each text instance and (b) the average labeling speed for each text instance, which is obtained from 10 annotators on average.
}
\label{fig:anno}
\end{figure}


In this paper, we research into the difficulty of scene text detection that how to use a simpler labeling method to achieve the same performance as the current state-of-the-art methods.
Similar to Wu~\emph{et al.}~\cite{texts_as_lines}, we adopt scribble lines to annotate texts in a coarse manner. As shown in Fig.\ref{fig:intro}, different from polygon annotations, these scribble lines are simply annotated with points, and roughly depict the text instances. Compared with the previous labeling methods, the proposed labeling method has the following advantages:
(1) It is a simpler labeling method with fewer point coordinates and faster labeling speed, as shown in Fig.~\ref{fig:anno}.
(2) It is a general labeling method for scene text detection because it can be applied to texts with various shapes, including horizontal texts, multi-oriented texts, and curved texts.
(3) Based on these scribble line annotations, our experiment results demonstrate that we can achieve the performance of the state-of-the-art methods.


To compensate for the loss of edge information from our weak annotations, we propose a new scene text detection framework, which can be trained in a weakly supervised manner.
Especially, an online proposal sampling method and a boundary reconstruction module are further introduced for character prediction and post-processing, respectively.
We conduct extensive experiments on several benchmarks to demonstrate the effectiveness of our scene text detection method, which can bridge the performance gap between our weakly labeling method and the original ones, with even better performance.

The contributions of this paper are three-fold.
\begin{itemize}
\item We propose a weakly labeling method for scene text detection with much lower labeling costs, which is a general method for texts with various shapes.
\item A weakly-supervised scene text detection framework is proposed for the weak annotations.
It outperforms the state-of-the-art methods that use original annotations on the Total-Text dataset and the CTW1500 dataset.
\item We will release our annotations and hope that it will benefit the research community in the field of weakly supervised scene text detection to achieve better performance with lower annotation costs.
\end{itemize}

\section{Related Work}

\subsection{Annotations for Scene Text Detection}
The traditional annotations for scene text detection can be roughly classified into three types. 
The first one is the axis-aligned bounding box that is mostly used to label the horizontal text datasets, such as the ICDAR 2013 dataset~\cite{ic13}. 
The second type is the quadrilateral or oriented rectangular bounding box for multi-oriented text datasets, such as the ICDAR 2015 dataset~\cite{ic15} and the MSRA-TD500 dataset~\cite{TD500}. 
The third type is the polygonal bounding boxes for arbitrary-shaped text datasets, such as the Total-text dataset~\cite{totaltext} and the CTW1500 dataset~\cite{ctw1500}.
The above-mentioned annotations may not include detailed shape information (e.g. the first two types) for text instances of irregular shapes or require heavy labeling costs (e.g. the third type).

Recently, Wu~\emph{et al.}~\cite{texts_as_lines} propose to draw continuous lines for text detection and use two coarse masks for text instances and background, respectively. However, continuous lines need more storage space than points and they are hard to draw without touch devices such as touch screens or touch pens. Thus, we propose a new line-level labeling method to overcome these shortcomings.

\subsection{Scene Text Detection Algorithms}

\subsubsection{Fully Supervised Text Detection}

The methods for scene text detection can be roughly divided into regression-based and segmentation-based methods. The regression-based methods are mainly inspired by general object detection methods~\cite{SSD, faster-rcnn}. TextBoxes++~\cite{textboxes++} proposes to predict the multi-oriented texts by quadrilateral regression.
EAST~\cite{EAST} and DDR~\cite{DDR} propose to directly regress the offsets for text instances in the pixel level.
The segmentation-based methods are mainly based on FCN~\cite{FCN} to predict text instance in the pixel level. PSE-Net~\cite{PSE-Net} proposes to segment text instances by progressively expanding kernels at different scales. Liao~\emph{et al.}~\cite{DB} propose a segmentation network with a differentiable binarization module, and improve on both accuracy and speed.

\subsubsection{Weakly Supervised Text Detection}
Due to the high annotation requirement for texts with various shapes, it is a difficulty that how to reduce the annotation costs without harming the detection performance.
WeText~\cite{WeText} proposes a weakly supervised learning technique for text detection with a small fully labeled dataset and a large unlabeled dataset.
CRAFT~\cite{CRAFT} proposes a text detection method by using characters and the affinity between characters, and it uses word-level annotations of real data to generate pseudo labels for fine-tuning.
Wu~\emph{et al.}~\cite{texts_as_lines} propose a segmentation-based network to use the proposed coarse masks for training, but there is a large gap between their performance and those using full masks.

\section{Methodology}

In this section, we first introduce our proposed weakly labeling method and then describe the proposed weakly supervised scene text detection method which performs text detection with the weak annotations.

\subsection{Weakly Labeling Method}


\begin{figure}[tb]
\centering
\includegraphics[width=1.0\linewidth]{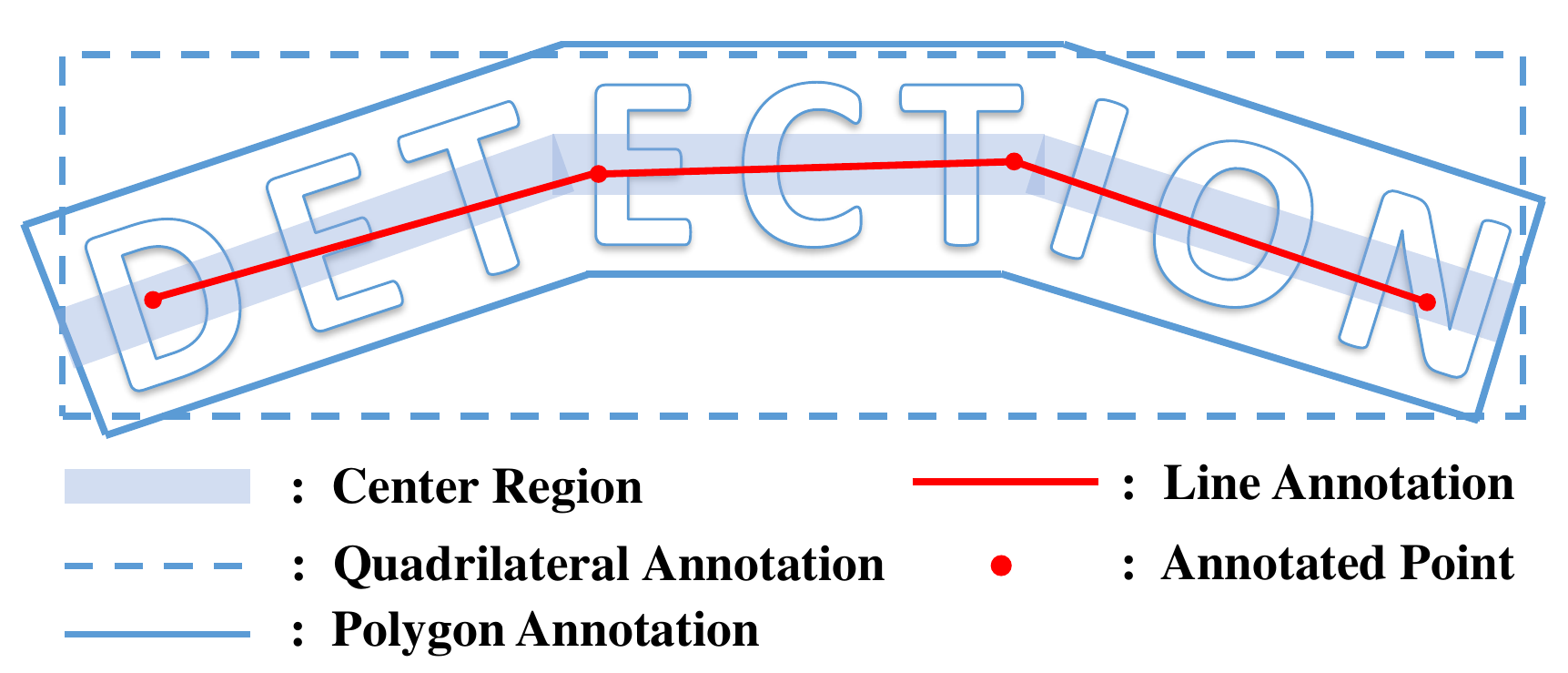}
\caption{The illustration of our proposed weakly labeling method. 
We annotate each text instance with a scribble line composed of several points.
}
\label{fig:labeling}
\end{figure}


Our proposed labeling method is annotating each text instance with a few points, as shown in Fig.~\ref{fig:labeling}. The labeling rule of a text instance is: 
(1) For horizontal or oriented text instances, two points near the centers of the first and the last characters are annotated;
(2) For irregular-shaped text instances, we annotate several points near the vertices of the center line.
The annotated points of a text instance are saved as a set of coordinates and can be connected into a scribble line. We keep the extreme blurry regions with the original annotation as they occupy only a small proportion and are not used in the training.



\subsection{Weakly Supervised Scene Text Detection}

\begin{figure*}[tb]
\centering
\includegraphics[width=1.0\linewidth]{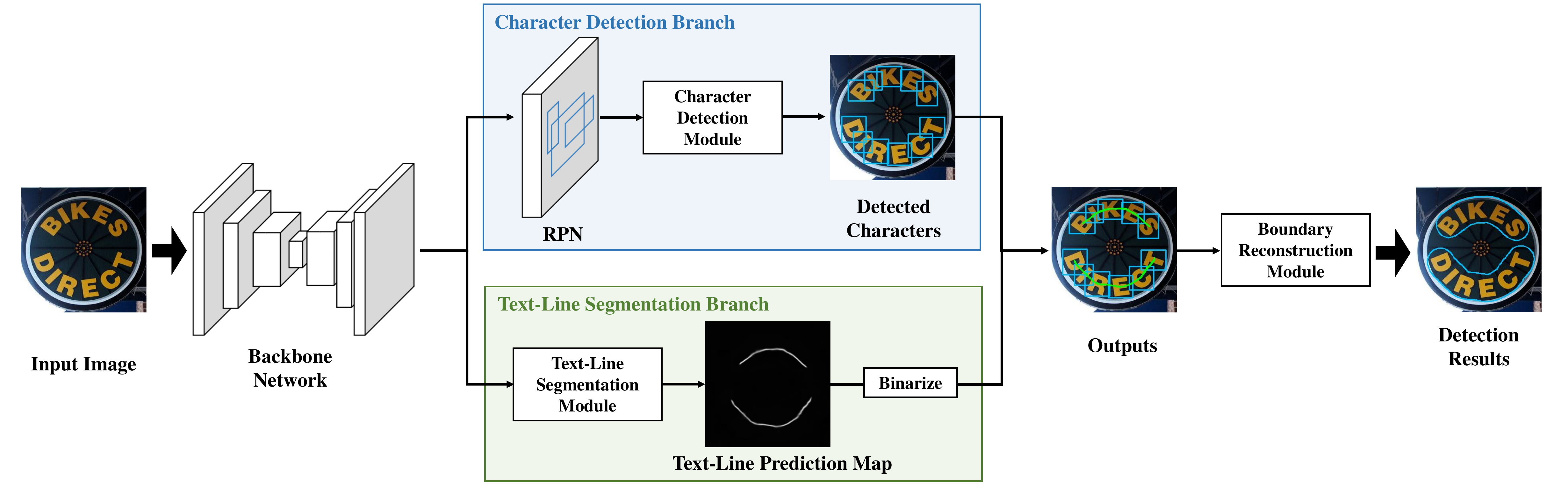}
\caption{The pipeline of our proposed method. The backbone network is with an FPN structure. The character detection branch and text-line segmentation branch are used to predict character boxes and text-lines. The final detection results are generated by the boundary reconstruction module.
}
\label{fig:pipeline}
\end{figure*}

Compared with multi-oriented quadrilaterals and polygons, the proposed weak annotations lose the edge information, which is indispensable to previous scene text detection methods. 
Hence, we propose a weakly supervised scene text detection framework, as shown in Fig.~\ref{fig:pipeline}. In the following parts of this section, we introduce the \textbf{Network Architecture}, \textbf{Pseudo Label Generation}, \textbf{Training Strategy}, \textbf{Optimization}, and \textbf{Inference}, respectively.

\subsubsection{Network Architecture}
The network consists of three parts, including a \textit{Backbone Network}, a \textit{Character Detection Branch}, and a \textit{Text-Line Segmentation Branch}.

The \textit{Backbone Network} consists of a ResNet-50~\cite{ResNet} backbone with deformable convolution~\cite{DCN} and a Feature Pyramid Network (FPN)~\cite{FPN} structure. The output feature maps are with 5 different scales (\emph{i.e.} $1/4$, $1/8$, $1/16$, $1/32$, $1/64$).

The \textit{Character Detection Branch} is a two-stage detection network, which consists of a Region Proposal Network (RPN)~\cite{faster-rcnn} and a character detection module. 
The RPN is used to predict the coarse character proposals.
The character detection module classifies the character proposals into different classes and regresses the offsets of the coarse proposals for refinement.

The \textit{Text-Line Segmentation Branch} predicts a probability map for the text-line location, which is supervised by the scribble line. The detailed network of the text-line segmentation branch is illustrated in the appendix.

\begin{figure}[tb]
\centering
\includegraphics[width=1.0\linewidth]{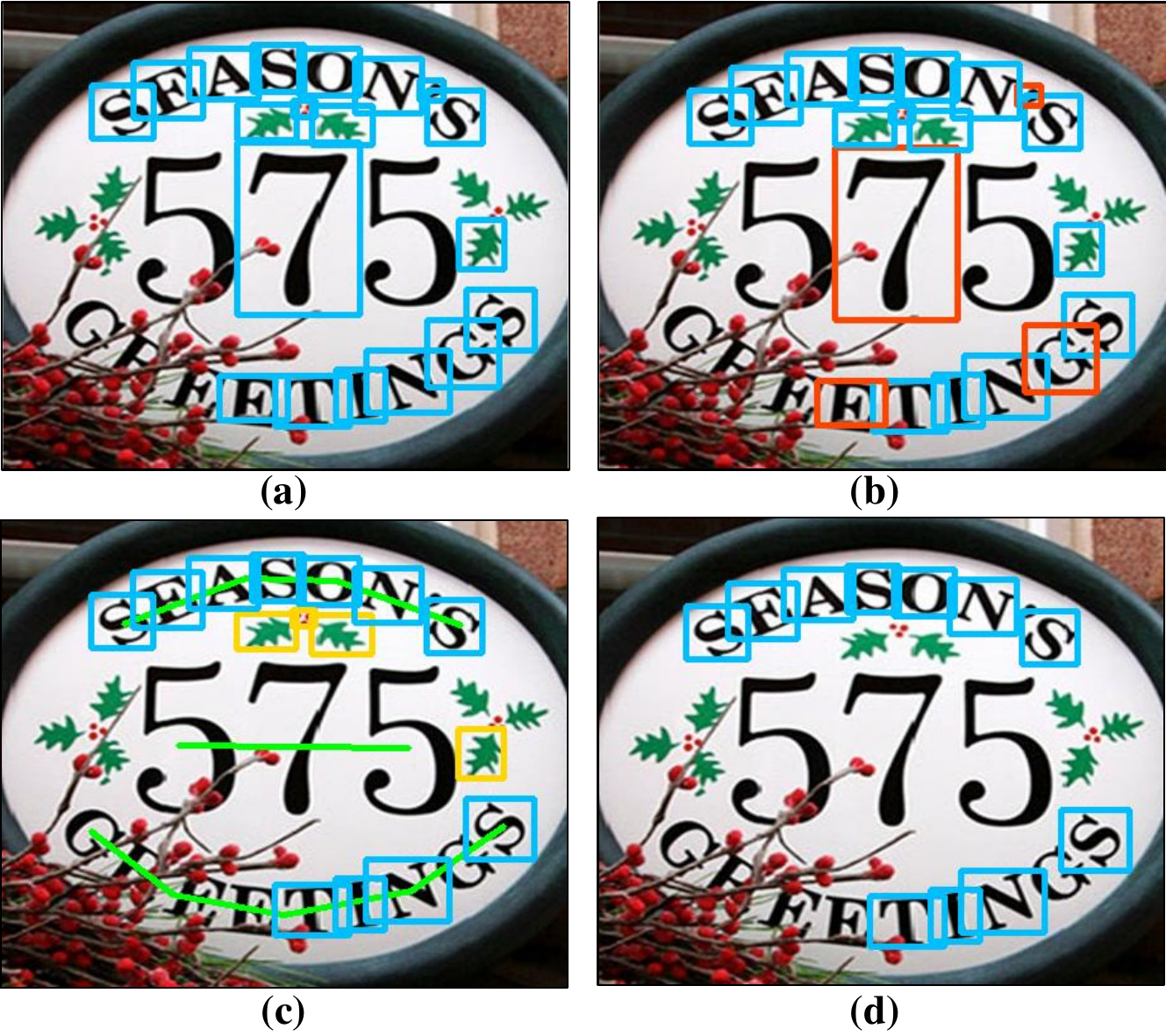}
\caption{ An example of the pseudo label generation: (a) Predict character boxes by a pre-trained model. (b) Remove boxes (red) with low scores and the ``Unknown" class. (c) Remove boxes (yellow) without overlap with any scribble lines. (d) The final pseudo labels for characters.
}
\label{fig:pseudo}
\end{figure}

\subsubsection{Pseudo Label Generation}
We propose to generate pseudo labels for the character detection branch from the model pre-trained with synthetic data~\cite{SynthText}, since the real data is only annotated with our weakly labeling method while the character-level annotations can be easily obtained from the synthetic data. 
As shown in Fig.~\ref{fig:pseudo}, the pseudo label generation consists of four steps:
(1) Use the pre-trained model to predict character boxes, scores, and classes of the real training data.
(2) Remove the character boxes with scores less than a threshold $T_{pseudo}$ or with the ``Unknown" class.
(3) Remove the character boxes without overlap with any scribble lines.
(4) The remained boxes are saved as the pseudo labels for training.

\subsubsection{Training Strategy}
The model is first pre-trained with synthetic data and then fine-tuned with both synthetic data and real data.

For the character detection branch, the RPN is only fine-tuned with synthetic data to avoid introducing the noise of wrong pseudo labels in the first stage of character detection. 
The second stage is fine-tuned with synthetic and real data with our online proposal sampling.
For real data, the positive samples are selected from the proposals matched with pseudo labels, and the negative samples are selected by the following two rules: (1) Those do not match any labels; (2) Those have no overlap with any scribble lines or difficult text instances; Then, the rest of the proposals are ignored in the training.
Due to the domain gap between synthetic data and real data, the pre-trained model can not generate pseudo labels for all characters as shown in Fig.~\ref{fig:pseudo}.
The proposals, which can not match any labels but have overlap with scribble lines, are the potential positive proposals. Hence, we should not regard these proposals as the negative, or it will have a negative influence on the detection performance.

The text-line segmentation branch is supervised with the scribble line ground-truths of synthetic data and real data.

\begin{algorithm}[tb]
  \caption{Boundary Reconstruction}
  \begin{algorithmic}[1]
    \Require
        Detected Character boxes $\mathbf{B}=\{B_i\}$ and scores $\mathbf{S}=\{S_i\}$;
        Detected Text-Lines $\mathbf{L}=\{L_j\}$;
    \Ensure
        Text detection results $\mathbf{T}$
    
    \For{each pair of $(B_i, S_i)$ in $\mathbf{B}$ and $\mathbf{S}$}
        \State If $S_i <$ a threshold $T_{infer}$: remove $B_i$ from $\mathbf{B}$
        \State If $B_i$ has the maximal overlap with $L_j$ in $\mathbf{L}$: put $B_i$ into group $G_j$
    \EndFor

    \For{All boxes $\mathbf{B}_j$ in each group $G_j$}
        \State $D = \frac{1}{N}\sum_{k=0}^{N-1} \sqrt{h_k * w_k}$, where $h_k$ and $w_k$ are the height and width of the box in $\mathbf{B}_j$.
        \State Expand the contour of the associated text-line $L_i$ to reconstruct the text boundary $T_i$ by a distance $D$
    \EndFor
    \State Output detection results $\mathbf{T} = \{T_i\}$
  \end{algorithmic}
  \label{code:Post-Processing}
\end{algorithm}

\subsubsection{Optimization}
The objective function consists of three parts, which is defined as follows.
\begin{equation}
    Loss = L_{rpn} + L_{char} + L_{line},
\end{equation}
where the $L_{rpn}$ is the loss of RPN which is identical to it in~\cite{faster-rcnn}, so we do not describe it in detail.

We apply a smooth L1 loss for the axis-aligned character box regression and a cross-entropy loss for character classification, which is defined as follows.
\begin{equation}
    \begin{aligned}
        L_{char} = L_{regress} + L_{classify} \\
         = L1_{smooth}\{\widehat{B}, B\} + L_{CE},
    \end{aligned}
\end{equation}
where $B$ refers to the predicted offsets from positive character proposals to associated ground-truths and $\widehat{B}$ is the target offsets, which are selected by our online proposal sampling.

The $L_{line}$ is the loss of the text-line segmentation module, and we adopt a binary cross-entropy loss for the text-line prediction map. Due to the imbalance between the number of text pixels and non-text pixels, we apply the online hard negative mining strategy proposed in~\cite{OHEM} to avoid the bias, and set the ratio of positive samples and negative samples as 1:3. It should be noted that the pixels in the difficult text instances will not be sampled.

We define $L_{line}$ as follows,
\begin{equation}
    L_{line} = \frac{1}{N}\sum_{i \in P}(1 - y_i)\log(1 - x_i) + y_i \log{x_i},
\end{equation}
where $P$ refers to the sampled set, and $N$ is the number of samples.

The text-line segmentation ground-truth is generated by connecting the annotated points of each text instance into a scribble line and draw it with $thickness=5$ on a map.

\subsubsection{Inference}
In the inference period, we apply a boundary reconstruction algorithm to generate accurate text boundaries from the predicted text scribble lines and the character boxes, as shown in Algorithm~\ref{code:Post-Processing}.

In the previous character-based text detection algorithms~\cite{seglink, CRAFT}, they propose to use the ``link" or ``affinity" to connect detected characters into the final detection results. 
However, noisy or missing characters probably lead to wrong or split detection results.
As shown in Algorithms.~\ref{code:Post-Processing}, the final detection results are generated by the geometry information of character boxes on average. It can suppress the impact of the noise, and will not suffer from the missing characters. Hence, our method is more robust compared with the previous weakly supervised methods and can achieve better performance.

\section{Experiment}

\subsection{Datasets}

The synthetic dataset is used for pre-training our model, and we fine-tune on each real dataset respectively. We use our weak annotations for training and the original annotations for evaluation.

\textbf{SynthText}~\cite{SynthText} is a synthetic dataset including 800k images. We use the no-cost character bounding boxes with classes and the text-lines composed of characters' center points.

\textbf{Total-Text}~\cite{totaltext} is a word-level based English text dataset. It consists of 1255 training images and 300 testing images, which contain horizontal texts, multi-oriented texts, and curved texts.

\textbf{CTW1500}~\cite{ctw1500} is a curved English text dataset that consists of 1000 training images and 500 testing images. All the text instances are annotated with 14 vertices.

\textbf{ICDAR2015}~\cite{ic15} contains natural images that are captured by Google Glasses casually, and most of them are severely distorted or blurred.
There are 1000 training images and 500 testing images, which are annotated with quadrilaterals.

\textbf{MSRA-TD500}~\cite{TD500} is a multi-lingual long text dataset for Chinese and English. It includes 300 training images and 200 testing images with arbitrary orientations. Following previous works~\cite{DB, TextSnake}, we include HUST-TR400~\cite{TR400} as training data in the fine-tuning.

\subsection{Implementation Details}
The training procedure can be roughly divided into 2 steps: 
(1) We pre-train our model on SynthText for 300k iterations.
(2) For each benchmark dataset, we use the pre-trained model to generate pseudo labels for characters, and fine-tune the model with pseudo labels and weak annotations for 300k. For each mini-batch, we sample the training images with a fixed ratio 1:1 for SynthText and the benchmark data.

The models are trained on two Tesla V100 GPUs with a batch size of 2.
We optimize our model by SGD with a weight decay of 0.0001 and a momentum of 0.9. The initial learning rate is set to 0.0025, and it is decayed by a factor 0.1 at iteration 100k and 200k. 
For training, the shorter sides of training images are randomly resized to different scales (\emph{i.e.} 600, 800, 1000, 1200). For inference, the shorter sides of the input images are usually resized to 1200. In both training and inference, the upper limit of the longer sides is 2333.
Moreover, we set the thresholds in our experiments as follows: (1) $T_{pseudo}$ for pseudo label generation is set to 0.9; (2) $T_{infer}$ for inference is usually set to 0.5;

\subsection{Comparisons with previous algorithms}

\begin{figure*}[tb]
\centering
\includegraphics[width=1.0\linewidth]{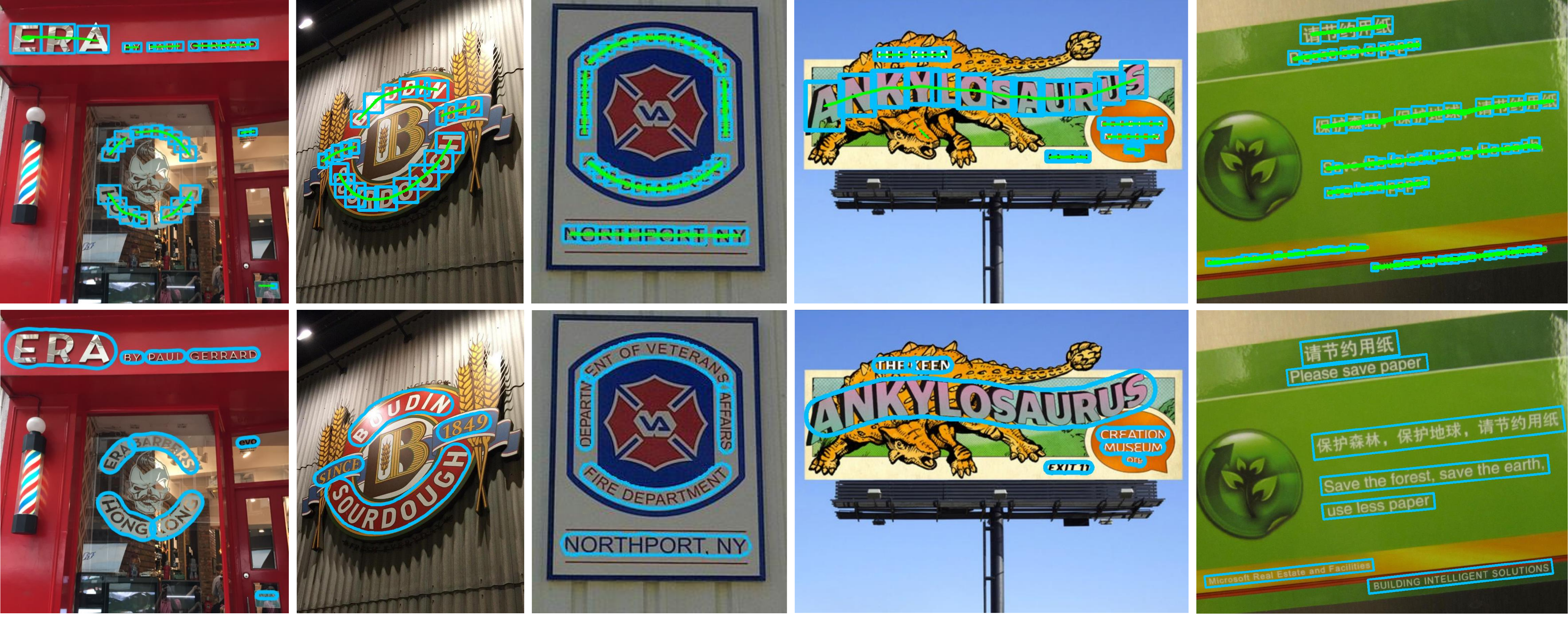}
\caption{The visualization of the network outputs and the final detection results in different cases, including horizontal texts, multi-oriented texts, curved texts, vertical texts, long texts, and multi-lingual texts.
The images in the first row show the detected characters and text-lines, and those in the second row show the text detection results after post-processing.
}
\label{fig:results}
\end{figure*}

To compare with previous fully supervised methods, we evaluate our scene text detection method on four standard benchmarks.
As shown in Fig.~\ref{fig:results}, we provide some qualitative results of texts in different cases, including horizontal texts, multi-oriented texts, curved texts, vertical texts, long texts, and multi-lingual texts.

\begin{table}[tb]
\centering
\begin{tabularx}{1.0\linewidth}{@{}l*{4}X@{}}
\toprule
Method                              & P             & R             & F   \\ \midrule
TextSnake~\cite{TextSnake}          & 82.7  & 74.5  & 78.4  \\
TextField~\cite{TextField}           & 81.2          & 79.9          & 80.6   \\
PSE-Net~\cite{PSE-Net}             & 84.0          & 78.0          & 80.9   \\
LOMO~\cite{LOMO} & 88.6 & 75.7  & 81.6  \\
ATRR~\cite{ATRR} & 80.9 & 76.2  & 78.5  \\
CRAFT~\cite{CRAFT}               & 87.6          & 79.9          & 83.6   \\
PAN~\cite{PAN}                 & 89.3          & 81.0          & 85.0   \\
DB~\cite{DB}                  & 87.1          & 82.5          & 84.7   \\
ContourNet~\cite{ContourNet}          & 86.9          & 83.9          & 85.4   \\
Zhang~\emph{et al.}~\cite{RRG-Net} & 86.5          & 84.9            & 85.7   \\
\midrule
*Texts as Lines~\cite{texts_as_lines} & 78.5 & 76.7 & 77.6 \\
*\textbf{Proposed Method}       & 89.7          & 83.5          & \textbf{86.5} \\ \bottomrule
\end{tabularx}
\caption{Detection results on the Total-Text dataset. ``P", ``R" and ``F" refer to precision, recall and f-measure. *using line-level weak annotations.}
\label{tab:total-text}
\end{table}

\begin{table}[tb]
\centering
\begin{tabularx}{1.0\linewidth}{@{}l*{4}X@{}}
\toprule
Method              & P             & R             & F   \\ \midrule
TLOC~\cite{ctw1500} & 77.4  & 69.8  & 73.4  \\
TextSnake~\cite{TextSnake}          & 67.9  & 85.3  & 75.6  \\
TextField~\cite{TextField}           & 83.0          & 79.8          & 81.4   \\
PSE-Net~\cite{PSE-Net}             & 84.8          & 79.7          & 82.2   \\
LOMO~\cite{LOMO}    & 89.2 & 69.6 & 78.4 \\
ATRR~\cite{ATRR}    & 80.1 & 80.2 & 80.1  \\
CRAFT~\cite{CRAFT}               & 86.0          & 81.1          & 83.5   \\
PAN~\cite{PAN}                 & 86.4          & 81.2          & 83.7   \\
DB~\cite{DB}                  & 86.9          & 80.2          & 83.4   \\
ContourNet~\cite{ContourNet}          & 84.1          & 83.7          & 83.9   \\
Zhang~\emph{et al.}~\cite{RRG-Net} & 85.9          & 83.0            & 84.5   \\
\midrule
*Texts as Lines~\cite{texts_as_lines} & 83.8 & 80.8 & 82.3 \\
*\textbf{Proposed Method}       & 87.2 & 85.0 & \textbf{86.1} \\ \bottomrule
\end{tabularx}
\caption{Detection results on the CTW1500 dataset. *using line-level weak annotations.}
\label{tab:CTW}
\end{table}

\subsubsection{Curved Text Detection}
As shown in Tab.~\ref{tab:total-text} and Tab.~\ref{tab:CTW}, we demonstrate the effectiveness of our method. 
On CTW1500 and Total-Text, our detector trained with our weak annotations outperforms the state-of-the-art fully supervised method by 1.6\% and 0.8\% on accuracy, respectively. 
The curved texts usually need higher annotation costs, but our weakly supervised method can achieve better performance with simpler annotations.
Especially, compared with~\cite{texts_as_lines}, our method improves a lot and bridges the performance gap between the weakly supervised methods and the fully supervised methods.

\subsubsection{Multi-Lingual Text Detection}
To evaluate our scene text detection method in other languages, we conduct an experiment on the MSRA-TD500 dataset. 
Due to the lack of Chinese texts in SynthText, we use a public code to generate ten thousand synthetic images with Chinese texts. After pre-training on both SynthText and our synthetic data, we fine-tune our model on the MSRA-TD500 dataset.
As shown in Tab.~\ref{tab:td500}, our method achieves a very close performance to the state-of-the-art methods.

\begin{table}[tb]
\centering
\begin{tabularx}{1.0\linewidth}{@{}l*{4}X@{}}
\toprule
Method              & P             & R             & F   \\ \midrule
EAST~\cite{EAST}    & 87.3  & 67.4  & 76.1  \\
TLOC~\cite{ctw1500-pr} & 84.5  & 77.1  & 80.6  \\
Corner~\cite{Corner}    & 87.6  & 76.2  & 81.5  \\
TextSnake~\cite{TextSnake}          & 83.2  & 73.9  & 78.3  \\
TextField~\cite{TextField}          & 87.4          & 75.9          & 81.3   \\
ATRR~\cite{ATRR}    & 85.2 & 82.1 & 83.6  \\
CRAFT~\cite{CRAFT}               & 88.2          & 78.2          & 82.9   \\
PAN~\cite{PAN}                 & 84.4          & 83.8          & 84.1   \\
DB~\cite{DB}                  & 91.5          & 79.2          & 84.9   \\
Zhang~\emph{et al.}~\cite{RRG-Net} & 88.1          & 82.3          & \textbf{85.1}   \\
\midrule
*Texts as Lines~\cite{texts_as_lines} & 80.6 & 74.1 & 77.2 \\
*\textbf{Proposed Method}       & 87.7          & 80.8          & 84.1 \\ \bottomrule
\end{tabularx}
\caption{Detection results on the MSRA-TD500 dataset. *using line-level weak annotations.}
\label{tab:td500}
\end{table}

\subsubsection{Multi-oriented Text Detection}
Most of the texts in ICDAR2015 are small, and some of them are severely distorted or blurred, which limits the performance of our character detection branch. 
Especially, our character labels of real data are generated by a model pre-trained on synthetic data, so it is much more difficult to generate accurate pseudo labels with the weak annotations. 
However, as shown in Tab.~\ref{tab:ic15}, our method still achieves a close performance compared with fully supervised methods, and outperforms the previous line-based weakly supervised method a lot.

\begin{table}[tb]
\centering
\begin{tabularx}{1.0\linewidth}{@{}l*{4}X@{}}
\toprule
Method              & P             & R             & F   \\ \midrule
DDR~\cite{DDR}  & 82.0  & 80.0  & 81.0  \\
EAST~\cite{EAST}    & 83.6  & 73.5  & 78.2  \\
Corner~\cite{Corner}    & 94.1  & 70.7  & 80.7  \\
TextSnake~\cite{TextSnake}          & 84.9  & 80.4  & 82.6  \\
TextField~\cite{TextField}          & 84.3          & 83.9          & 84.1   \\
PSE-Net~\cite{PSE-Net}             & 86.9          & 84.5          & 85.7   \\
LOMO~\cite{LOMO} & 91.3  & 83.5  & 87.2  \\
ATRR~\cite{ATRR}    & 89.2 & 86.0 & 87.6  \\
CRAFT~\cite{CRAFT}               & 89.8          & 84.3          & 86.9   \\
PAN~\cite{PAN}                 & 84.0          & 81.9          & 82.9   \\
DB~\cite{DB}                  & 91.8          & 83.2          & \textbf{87.3}   \\
ContourNet~\cite{ContourNet}          & 86.1          & 87.6          & 86.9   \\
Zhang~\emph{et al.}~\cite{RRG-Net} & 88.5          & 84.7          & 86.6   \\
\midrule
*Texts as Lines~\cite{texts_as_lines} & 81.7 & 77.1 & 79.4 \\
*\textbf{Proposed Method}       & 83.1          & 85.7          & 84.4 \\ \bottomrule
\end{tabularx}
\caption{Detection results on the ICDAR2015 dataset. *using line-level weak annotations.}
\label{tab:ic15}
\end{table}

\subsection{Annotation Deviation}

\begin{figure}[tb]
\centering
\includegraphics[width=1.0\linewidth]{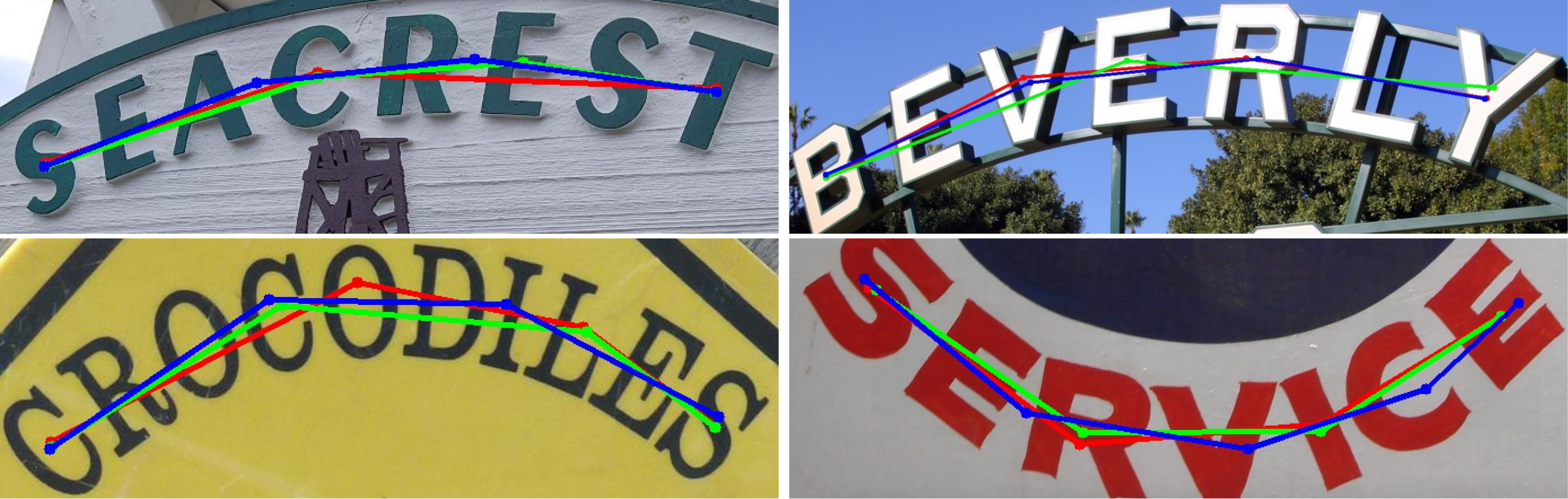}
\caption{The illustration of the annotations by different annotators (in different colors).}
\label{fig:difference}
\end{figure}

\begin{table}[tb]
\centering
\begin{tabularx}{1.0\linewidth}{@{}l*{4}X@{}}
\toprule
Annotation              & P             & R             & F   \\ \midrule
Annotation-1            & 89.8          & 82.6          & 86.0 \\
Annotation-2            & 88.2          & 83.7          & 85.9 \\
Annotation-3            & 89.7          & 83.5          & 86.5 \\ \bottomrule
\end{tabularx}
\caption{Detection results on the Total-Text dataset with different annotations by different annotators.}
\label{tab:annotation}
\end{table}

\begin{figure}[tb]
\centering
\includegraphics[width=1.0\linewidth]{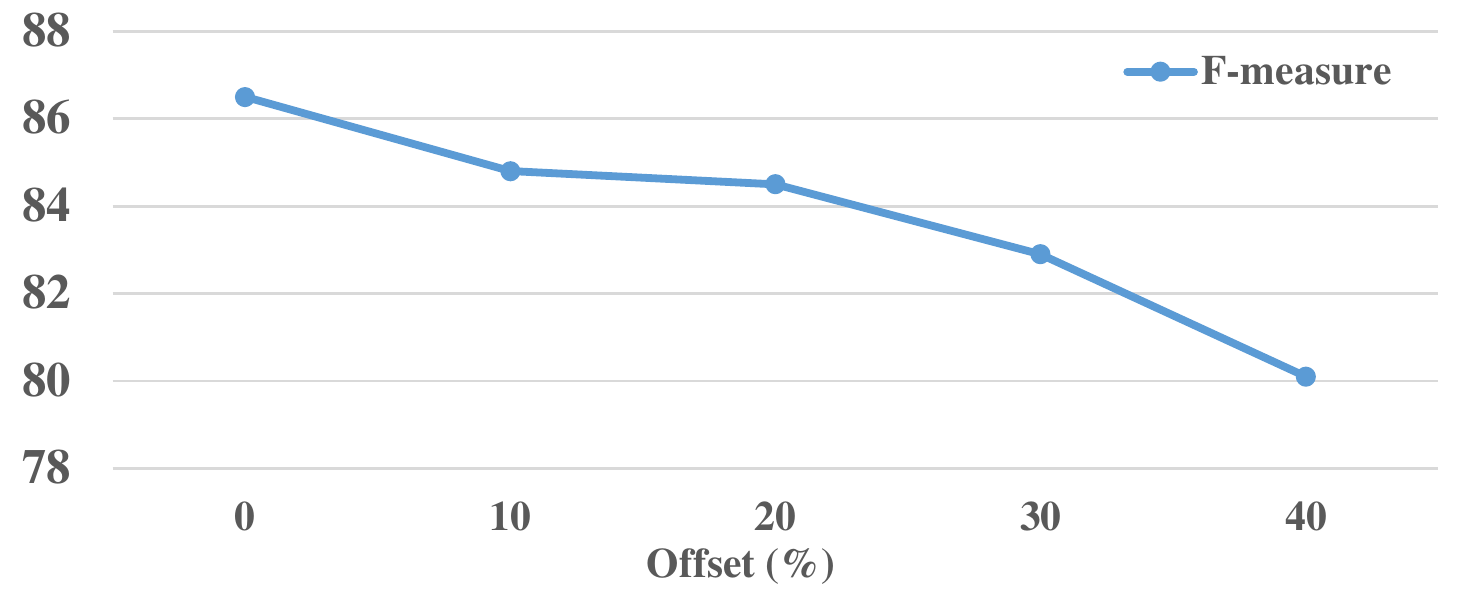}
\caption{Detection results on the Total-Text dataset with different manual noise, achieved by applying random offsets to the annotated coordinates.}
\label{fig:noise}
\end{figure}

Our proposed labeling method is annotating each text instance with a scribble line, and different annotators tend to annotate different point coordinates by their intuitions. 
Hence, we arrange different annotators to label the training data individually, and it is easy to find the differences among different sets of weak annotations as shown in Fig.~\ref{fig:difference}.
To show the robustness of our scene text detection method to these different annotations, we conduct experiments on Total-Text.
As shown in Tab.~\ref{tab:annotation}, the models trained with different annotations consistently achieve the state-of-the-art performance on the Total-Text dataset.

Furthermore, we add random noise to the Annotation-3 to study the influence of annotation deviation on the detection performance. 
For each point coordinate $(x, y)$, we randomly modify it into $(x \pm R_x * H, y \pm R_y * H)$, where $R_x$ and $R_y$ are randomly selected in $[-\frac{1}{2}\text{Offset}, +\frac{1}{2}\text{Offset}]$ and $H$ is the height of each text instance.
Comparing Fig.~\ref{fig:noise} and Tab.~\ref{tab:annotation}, we find that the deviation from different annotators has less negative influence than the random noise of 10\% offset, and the detection performance is still over 80\% f-measure when the annotations are seriously disturbed by 40\% offset.

\subsection{Discussion}

\subsubsection{Multi-Class Character Detection}

As shown in Tab.~\ref{tab:mc}, we study the character classification by three different settings and find that ``All+BF" achieves the best performance. Since the pseudo labels of character classes generated by the pre-trained model are not as accurate as the real labels in synthetic data, we only use ``Background" and ``Foreground" classification for real data instead of ``All" classes classification, which makes the model easier to optimize.
Moreover, it is easy to change our method into an end-to-end method in the second manner. We have directly obtained the recognition results by sorting the characters from left to right of each text instance, but we can only achieve 67.0 f-measure on Total-Text, which is far from the state-of-the-art end-to-end method.
In the future work, it needs the recognition annotations and better post-processing to achieve better performance for the text spotting task.

\begin{table}[tb]
\centering
\begin{tabularx}{1.0\linewidth}{@{}l*{4}X@{}}
\toprule
Dataset     & MC            & P             & R             & F   \\ \midrule
\multirow{3}{*}{Total-Text}  &  BF    & 88.4   & 79.5   & 83.7 \\
                             &  All    & 89.3   & 81.5   & 85.2 \\
                             &  All+BF    & 89.7   & 83.5   & \textbf{86.5} \\
\midrule
\multirow{3}{*}{CTW1500}     &  BF    & 87.6 & 80.8 & 84.1 \\
                             &  All    & 86.9 & 83.8 & 85.3 \\
                             &  All+BF    & 87.2 & 85.0 & \textbf{86.1} \\
\bottomrule
\end{tabularx}
\caption{Detection results on the Total-Text dataset and the CTW1500 dataset with different multi-class classification (short for ``MC") manners. ``BF": The character classes include ``Background" and ``Foreground"; ``All": The character classes for both synthetic and real data include ``Background", 10 numbers, 26 letters, and ``Unknown"; ``All+BF": The character classes for synthetic data include ``Background", 10 numbers, 26 lowercase letters, and ``Unknown", and those for real data only include ``Background" and ``Foreground".}
\label{tab:mc}
\end{table}

\subsubsection{Comparison with line-based text detection methods}
Compared with Texts as Lines~\cite{texts_as_lines}, our method has improved in the following aspects:

(1) Their weak annotations contain two images of coarse masks for texts and background, while our annotations only consist of point coordinates for each scribble lines. As a result, our weakly labeling method can save on both the annotation costs and the storage space.

(2) Their text detection method ignores the character information and has a big performance gap with state-of-the-art methods. 
Although our proposed annotations lose the edge information of text instances, our character detection branch and the boundary reconstruction module can compensate for the weakness with the geometric information of detected characters.
Demonstrated by our experiments, our method improves a lot on accuracy, and even outperforms state-of-the-art methods on the Total-Text dataset and the CTW1500 dataset.

\subsubsection{Limitation}
For the character detection branch, it is difficult to detect the distorted or small characters, which is the reason for the unremarkable performance on the ICDAR 2015 dataset. However, due to our proposed post-processing, there is no need to detect all the characters in a text instance, which reduces the negative influence of the disadvantages to some extent.

\section{Conclusion}

In this paper, we research into the difficulty that how to use a simpler labeling method for scene text detection without harming the detection performance.
We propose to annotate each text instance with a scribble line in order to reduce the annotation costs, while it is a general labeling method for texts with various shapes.
Moreover, we present a weakly supervised framework for scene text detection, which can be trained with our weak annotations of real data and no-cost annotations of synthetic data. 
By our experiments on several benchmarks, we demonstrate that it is possible to achieve state-of-the-art performance based on the weak annotations, even better.
For research purposes, we will release the annotations of different benchmarks in our experiments, and we hope it will benefit the field of scene text detection to achieve better performance with simpler annotations.

{
\bibliography{reference}
}
\end{document}


\linenumbers %
\maketitle

\section{Network Architecture}
Our network is composed of three parts, including a Backbone Network, a Character Detection Branch and a Text-Line Segmentation Branch. 
The network implementation is based on the official code\footnote{https://github.com/facebookresearch/maskrcnn-benchmark} of Mask-RCNN~\cite{mask-rcnn}.

\subsection{Backbone Network}
The Backbone Network consists of a ResNet-50~\cite{ResNet} backbone with deformable convolution~\cite{DCN}, and a Feature Pyramid Network (FPN)~\cite{FPN} structure. The detailed configurations of the ResNet-50 is shown in Tab.~\ref{fig:configuration}.
The different feature maps from ResNet-50 are fed into the FPN structure which is implemented based on the official code of Mask-RCNN.
The final output feature maps are with 5 different scales (\emph{i.e.} $1/4$, $1/8$, $1/16$, $1/32$, $1/64$).

\begin{table}[tb]
\centering
\includegraphics[width=1.0\linewidth]{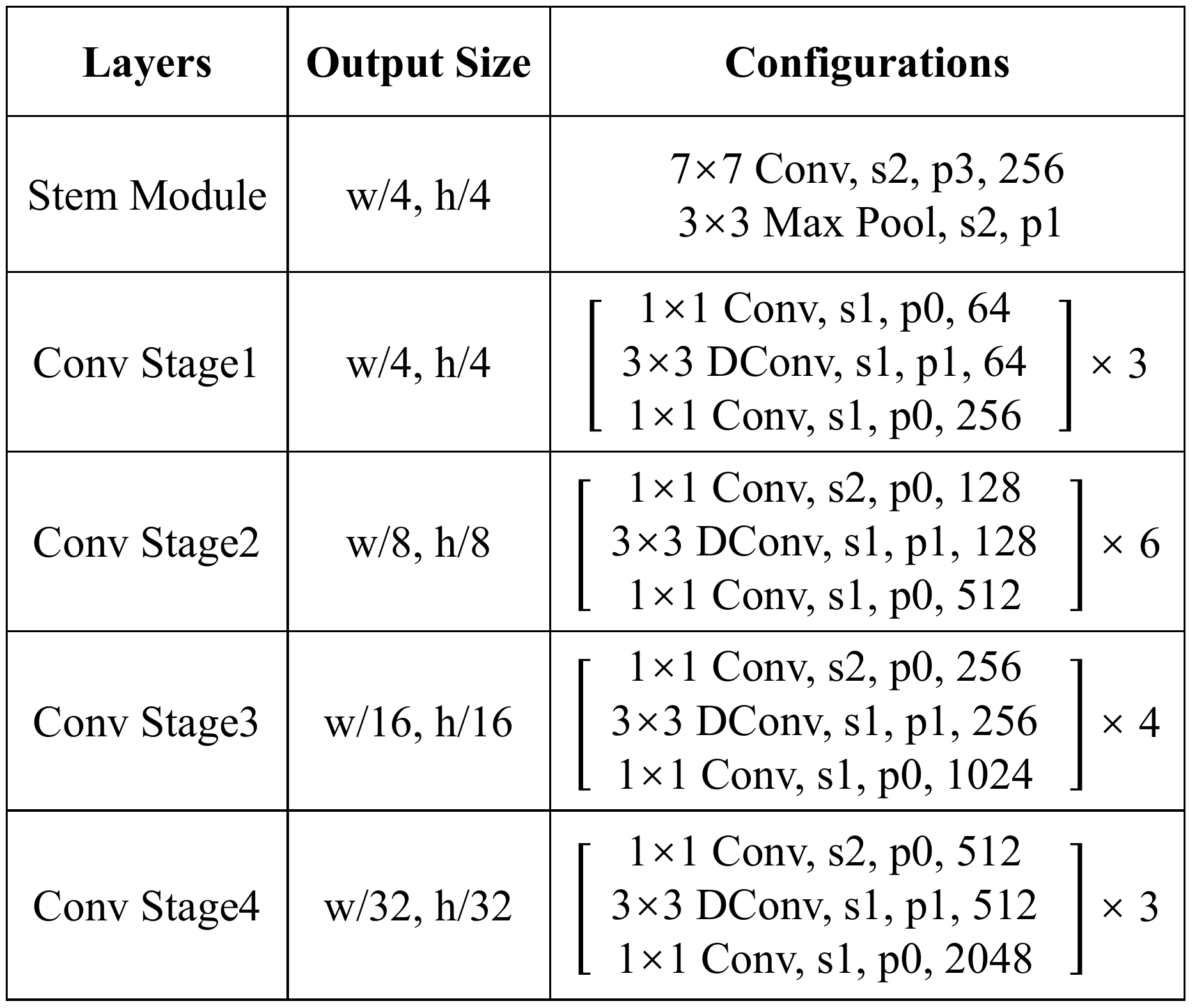}
\caption{The detailed configurations of ResNet-50 with deformable convolution. ``7$\times$7 Conv, s2, p3, 256" means a convolution layer with kernel size = 7, stride = 2, paddding = 3 and output channel = 256. ``DConv" means the deformable convolution layer.
}
\label{fig:configuration}
\end{table}

\subsection{Character Detection Branch}
\begin{figure*}[tb]
\centering
\includegraphics[width=1.0\linewidth]{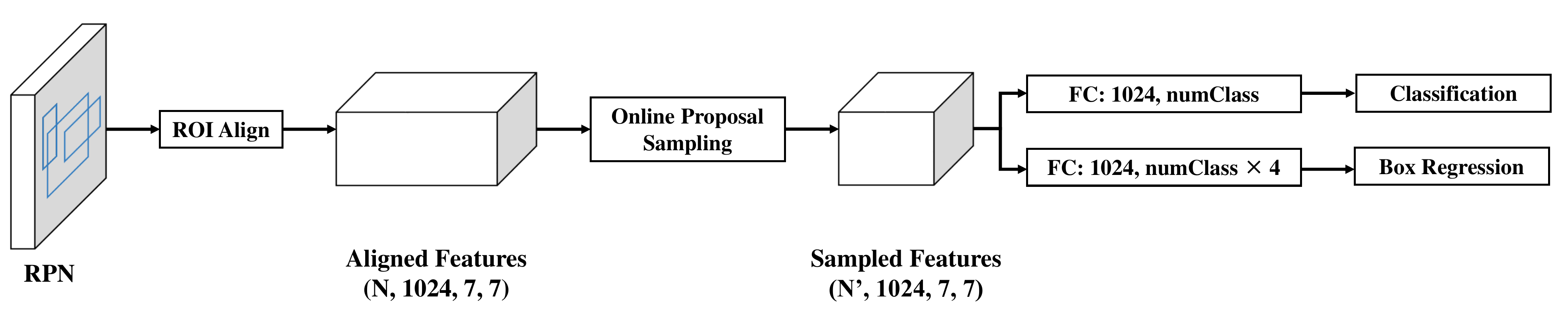}
\caption{The detailed structure of our character detection branch. We propose the online proposal sampling to better select samples in the weakly supervised learning. The sampled features are fed into two fully connected layers for refinement and classification.
}
\label{fig:char}
\end{figure*}

As shown in Fig.~\ref{fig:char}, the Character Detection Branch is a two-stage detection network, which consists of a Region Proposal Network (RPN)~\cite{faster-rcnn} for coarse character proposals predicting and a character detection module for proposals refinement.
We take the coarse proposals from the RPN and select positive and negative proposals via our online proposal sampling. The features of different coarse proposals are extracted by the ROI-Align technique~\cite{mask-rcnn}.
Then, we use two fully connected layers to regress the offsets of the coarse proposals for character refinement and classify the characters into different classes.

\subsection{Text-Line Segmentation Branch}
\begin{figure}[tb]
\centering
\includegraphics[width=1.0\linewidth]{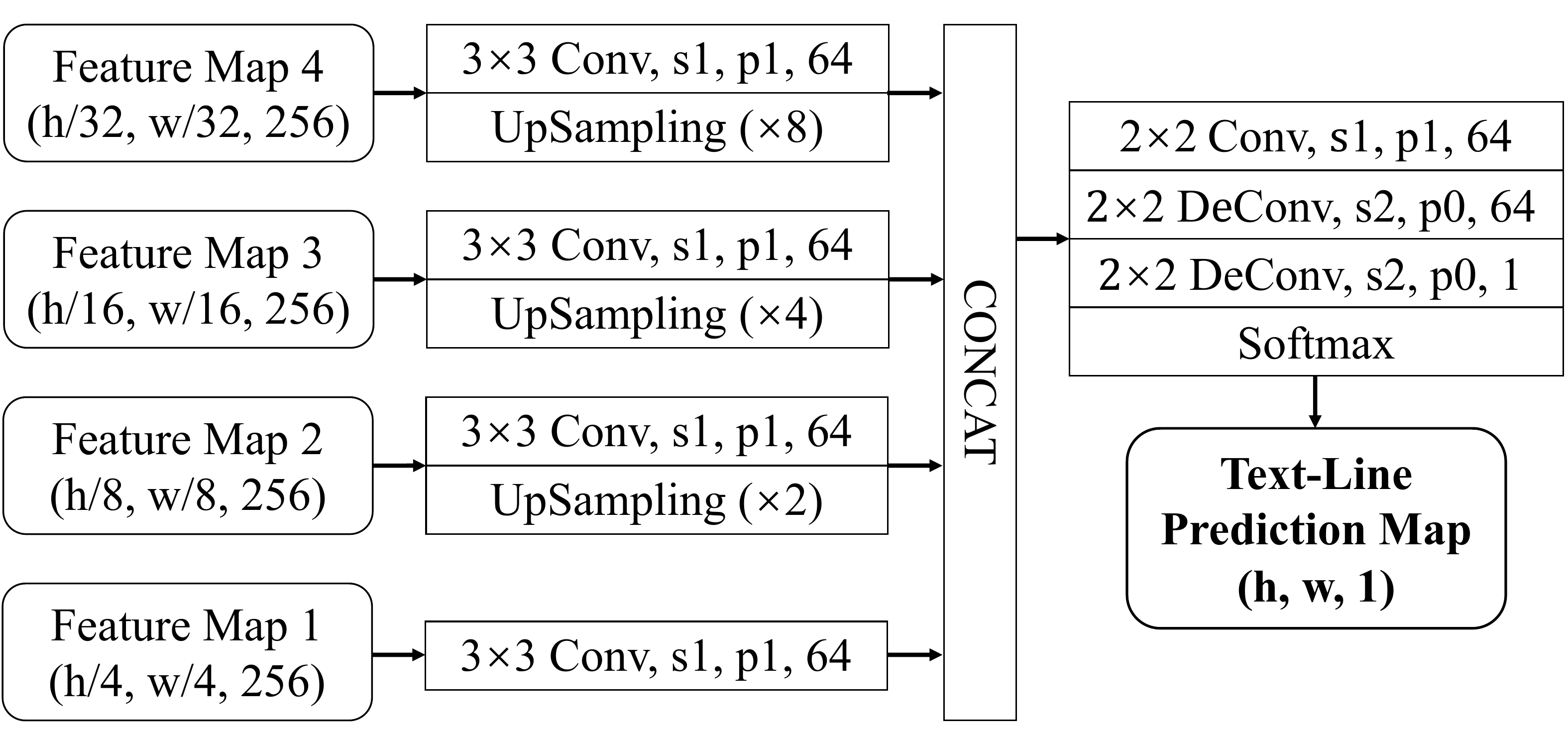}
\caption{The architecture of our text-line segmentation branch. ``3$\times$3 Conv, s1, p1, 64" means a convolution layer with kernel size = 3, stride = 1, padding = 1 and output channel = 64. ``DeConv" means the transposed convolution layer.
}
\label{fig:segmentation}
\end{figure}

The Text-Line Segmentation Branch predicts a probability map for the text-line location. The detailed architecture of the branch is illustrated in Fig.~\ref{fig:segmentation}. 
The Text-Line Segmentation Branch takes the four feature maps from the Backbone Network, except for the feature map with the scale of $1/64$, which is too small for text-line segmentation.
The feature maps in different scales are fed into a fully convolutional module with up-sampling. 
All feature maps are up-sampled by Nearest Neighbor interpolation to quarter size of the input image and concatenated in the channel-wise.
Then, the concatenated feature map is fed into a convolution layer and two transposed convolution layers, to predict a probability map for the text-line location.

\section{Implementation Details}
\subsection{Training Data}
We use the original training images from all the benchmarks, and the proposed annotations will be released after publication. It should be noted that we adopt a public code\footnote{https://github.com/JarveeLee/SynthText\_Chinese\_version} to generate ten thousand synthetic data with Chinese for the experiment of MSRA-TD500.

\subsection{Computing Infracture}
All the models are trained on two Tesla V100 GPUs with a batch size of 2, and the operating system is Centos 7.
The main software libraries are listed as follows:
(1) GCC: 5.3.1
(2) CUDA: 9.0
(3) PyTorch: 1.1.0
(4) torchvision: 0.3.0

\subsection{Training Period}
The training procedure can be roughly divided into 2 steps: 
(1) We pre-train our model on SynthText for 300k iterations.
(2) For each benchmark dataset, we use the pre-trained model to generate pseudo labels for characters, and fine-tune the model with pseudo labels and weak annotations for 300k. For each mini-batch, we sample the training images with a fixed ratio 1:1 for SynthText and the benchmark data.

\subsection{Hyper-parameter Setting}
We optimize our model by SGD with a weight decay of 0.0001 and a momentum of 0.9. The initial learning rate is set to 0.0025, and it is decayed by a factor 0.1 at iteration 100k and 200k. 
For training, the shorter sides of training images are randomly resized to different scales (\emph{i.e.} 600, 800, 1000, 1200). For inference, the shorter sides of the input images are resized to 1200 for the Total-Text, ICDAR 2015 and CTW1500 datasets, and 800 for the MSRA-TD500 dataset. In both training and inference, the upper limit of the longer sides is 2333.
Moreover, we set some thresholds in our experiments as follows: 
(1) $T_{pseudo}$ for pseudo label generation is set to 0.9; 
(2) $T_{infer}$ for inference is set to 0.5 for the Total-Text, ICDAR 2015 and CTW1500 datasets, and 0.7 for the MSRA-TD500 dataset.
(3) The threshold for binarization in the text-line segmentation branch is set differently according to different benchmarks. It is better to choose it from $[0.1, 0.3]$ for the Total-Text, ICDAR 2015, and MSRA-TD500 datasets, while it should be set to 0.03 or 0.04 for the CTW1500 dataset;
More details can be referred to our code after publication.

\subsection{Random Seed}
We do not fix the random seeds in our experiments, so it is not easy to provide any numbers. However, we have trained our models with different random seeds for many times, and the randomness has a small influence on the final detection performance.

\subsection{Evaluation Metric}
Following the previous work in the field of scene text detection, we evaluate our detection results by precision, recall, and f-measure as shown below.
\begin{equation}
    \text{Precision} = \frac{\text{Matched Detection Results}}{\text{All Detection Results}}
\end{equation}
\begin{equation}
    \text{Recall} = \frac{\text{Matched Detection Results}}{\text{All Ground-Truths}}
\end{equation}
\begin{equation}
    \text{F-measure} = \frac{2*\text{Precision}*\text{Recall}}{\text{Precision} + \text{Recall}}
\end{equation}

{
\bibliography{reference}
}